\documentclass{article}
\usepackage[final]{colm2026_conference} %
\usepackage{microtype}
\usepackage{booktabs}
\usepackage{longtable}
\usepackage{array}
\usepackage{calc}
\usepackage{amsmath}
\usepackage{amssymb}
\usepackage{graphicx}
\usepackage[most]{tcolorbox}
\tcbuselibrary{breakable}
\newenvironment{promptbox}[1]{%
  \begin{tcolorbox}[colback=gray!5!white, colframe=gray!50!black, fonttitle=\bfseries\small,
    title={#1}, breakable, left=4pt, right=4pt, top=4pt, bottom=4pt,
    before upper={\setlength{\parskip}{2pt}}]\ttfamily\footnotesize}{\end{tcolorbox}}
\definecolor{binirr}{HTML}{B2182B}
\definecolor{binrec}{HTML}{2166AC}
\definecolor{binres}{HTML}{595959}
\newtcolorbox{casebox}[2]{colback=#1!6!white, colframe=#1!65!black, coltitle=white,
  fonttitle=\bfseries\footnotesize, title={#2}, left=4pt, right=4pt, top=2pt, bottom=2pt,
  boxsep=2pt, fontupper=\footnotesize, before upper={\setlength{\parskip}{1pt}}}
\usepackage{hyperref}
\usepackage{url}
\providecommand{\tightlist}{\setlength{\itemsep}{0pt}\setlength{\parskip}{0pt}}

\definecolor{darkblue}{rgb}{0, 0, 0.5}
\hypersetup{colorlinks=true, citecolor=darkblue, linkcolor=darkblue, urlcolor=darkblue}

\title{What Eviction Destroys: A Restore-Counterfactual Audit of\\ Forgetting in Agent Memory}
\author{Chen Shen \\ Megagon Labs \\ \texttt{chen\_s@megagon.ai}}

\begin{document}
\maketitle
\lhead{Accepted at the COLM 2026 Workshop on Context Beyond the Window}

\begin{abstract}
Agent memory systems must discard stored information when their history exceeds a fixed token budget. Existing budget--accuracy frontiers quantify the resulting loss in accuracy, but do not distinguish irreversible losses caused by eviction from recoverable retrieval failures. We introduce the restore counterfactual, a per-question paired intervention that reinstates the question's gold evidence in the read-time context and reruns the same reader. Combining the change in correctness with whether the evidence was retained after eviction classifies each oracle-answerable error as recoverable, irreversible, or residual; in the residual case, the answer remains incorrect after restoration. We evaluate FIFO, random, redundancy-aware, and LLM-importance eviction on LongMemEval-S at three budgets and under two retrieval regimes, using GPT-4o-mini as the primary reader and judge and GPT-5.4-mini as a robustness reader. Under top-\(k\) retrieval at an 80k-token budget, the irreversible share among errors corrected by restoration is 0.67--0.73 for FIFO, random, and redundancy-aware eviction, compared with 0.60 for LLM-importance. At 8k tokens, it reaches 1.00 for all four policies. Recoverable errors occur under top-\(k\) retrieval at 80k tokens but are absent under forced-gold injection by construction, so budget--accuracy results are not directly comparable unless the retrieval regime is reported. An exploratory matched-accuracy analysis detects no difference in irreversible rate among accuracy-matched policy pairs at a resolution of 1.2--6 percentage points. The same analysis detects the deliberately destructive control. To our knowledge, this is the first per-item, per-question restore-counterfactual audit of eviction for external agent-memory stores on a standard conversational benchmark.
\end{abstract}

\section{Introduction}\label{introduction}

In multi-session interaction, an LLM agent accumulates more experience than fits its context window. Such agents use a memory pipeline that writes information, manages what is retained or forgotten, and reads information \citep{zhangMemorySurvey2024,sumersCoALA2024}. Systems \citep{parkGenerativeAgents2023,kangMemoryOS2025,chhikaraMem02025} and benchmarks \citep{maharanaLoCoMo2024,wuLongMemEval2024} make this pipeline concrete. Recent work on memory budgets includes theoretical \citep{zouRateDistortionMem2026} and empirical \citep{zhangBudgetMem2026} budget--accuracy frontiers, as well as learned retention policies \citep{liEMBER2026,kangOSLMR2026}. These studies quantify the accuracy lost under tighter budgets, but do not characterize the composition of that loss.

A frontier point conflates two failure modes with opposite remedies. If the needed evidence was evicted from
the store, no retriever recovers it and only more retention helps (the loss is \emph{irreversible}); if it
survived but was not retrieved, a better retriever helps with no extra retention (\emph{recoverable}). An
all-recoverable frontier and an all-irreversible frontier can look identical on the accuracy axis but require
different interventions.

The \emph{restore counterfactual} starts from a policy's post-eviction store, reinstates the question's gold evidence
in the read-time context, reruns the same frozen reader, and records any change in correctness. The change in
correctness, together with whether the gold evidence was retained after eviction, classifies each
oracle-answerable policy error as recoverable, irreversible, or residual; in the residual case, the answer remains incorrect after
restoration. We contribute the instrument and the decomposition. We do not propose a new memory system or rank
policies. Each outcome has a different diagnostic interpretation: a large irreversible share points to
retention; a large recoverable share points to retrieval; and a large residual share indicates that the reader
failed to use the restored evidence. The same decomposition applies when forgetting is required by privacy or
data-protection constraints; it measures whether the resulting loss in task accuracy is recoverable or
irreversible.

\textbf{Contributions.} (i) The restore-counterfactual metric for eviction in external agent-memory stores (\S{}3.1);
(ii) the recoverable/irreversible/residual decomposition on an oracle-answerable denominator (\S{}3.2); and (iii) a
controlled LongMemEval-S study of four eviction policies (including an LLM-importance arm) across three budgets
and two retrieval regimes (\S{}4--\S{}5). The study shows that budget--accuracy frontiers are not directly comparable
unless the retrieval regime is reported.

\section{Related work}\label{related-work}

\textbf{Budgeted memory and forgetting.} MaRS/FiFA compares eviction on synthetic generative-agent simulations
with a composite utility bound, not downstream accuracy \citep{alqithamiForgetfulFaithful2025}. EMBER \citep{liEMBER2026} and OSL-MR \citep{kangOSLMR2026} propose methods that learn budgeted retention
and report end-task accuracy. Neither distinguishes irreversible losses from recoverable ones. DeMem
\citep{zouRateDistortionMem2026} formalizes an optimal forgetting boundary; we give an operational
per-question estimator of whether it has been crossed. BudgetMem \citep{zhangBudgetMem2026} reports
accuracy--cost frontiers on LoCoMo and LongMemEval; our decomposition characterizes the loss at a frontier point
by failure type.

\textbf{Stage decomposition.} WhenLoss \citep{yuWhenLoss2026} decomposes memory failures into write- vs retrieval-side mass in aggregate.
Our decomposition uses a per-question intervention to attribute errors to eviction.
Observational diagnostics \citep{gargMemFail2026} and retrieval-vs-utilization splits
\citep{yuanDiagnosingMemory2026} locate or attribute failures, but none attributes them to a capacity-bounded
eviction decision via a per-item restore. The most closely related work, AgingBench \citep{zhuAgingBench2026}, uses
paired oracle-injection probes over an agent-written store to diagnose write/retrieval/utilization failures,
but it targets aging mechanisms on custom scenarios rather than capacity-bounded eviction policies. Here we
classify each oracle-answerable policy error by a per-question restore of the needed evidence on LongMemEval-S.

\textbf{The intervention class.} Reinstating an item and measuring the resulting change is a form of leave-one-out
context attribution \citep{cohenWangContextCite2024} and an inference-time analog of counterfactual memorization
\citep{zhangCounterfactualMem2023}; per-item downstream utility also appears in retrieval evaluation
\citep{salemiERAG2024}. To our knowledge, this is the first study to apply this class of interventions to
eviction from an external agent-memory store, with end-task accuracy as the outcome. That setting distinguishes
our audit from prior applications of such interventions and from oracle-injection diagnostics
\citep{zhuAgingBench2026}.

Context-paging employs restores as a serving mechanism
\citep{masonDemandPaging2026}. Because KV-cache eviction can outperform the full cache, an aggregate no-evict delta
mismeasures eviction damage \citep{buiMakeEachToken2026}. Systems that assume recoverability or repair
retrieval internally \citep{hsuHORMA2026,luREAL2026,semenovCWL2026} do not measure irreversible task loss caused
by eviction.

\textbf{Retrieval confound.} Because a frozen retriever can miss evidence retained in the store
\citep{derehagSmartSearch2026,wuLongMemEval2024}, eviction effects must be interpreted under controlled retrieval
conditions. We therefore report two retrieval settings and the difference between them as a result.

\section{The restore counterfactual}\label{the-restore-counterfactual}

Following the causal-measurement template \citep{lesciMemorisationProfiles2024}, we define the target quantity as the loss in end-task accuracy caused by a forgetting decision, identify it from observables under stated assumptions, and give the estimator.
\S{}3.1 defines the per-question estimator, and \S{}3.2 uses it to construct the three-bin decomposition.
\S{}3.3 states the assumptions needed to interpret the bins as irreversible, recoverable, and residual errors.

\subsection{The restore-counterfactual metric}\label{the-restore-counterfactual-metric}

\textbf{Why a counterfactual, not an accuracy drop.} A naive estimator measures the drop in accuracy after eviction. That drop conflates the two failure modes we need to separate: the evidence a question needs
may have been destroyed by eviction, or it may have survived in the store and merely gone unretrieved. The
two have opposite remedies yet are indistinguishable on the accuracy axis.
The restore counterfactual distinguishes these failure modes by reinstating the specific evidence the
question needs and checking whether the answer becomes correct. This is a form of leave-one-out context attribution \citep{cohenWangContextCite2024}.

\textbf{The estimator.} A question \(q\) is answered by a frozen reader over memory units injected from a
capacity-bounded store. For an eviction policy \(P\) at budget \(B\), let the post-eviction store be \(S_P\). The
restore counterfactual reinstates the question's full gold evidence set \(G_q\) (the units the benchmark labels
as containing the answer) into the read-time context and re-answers:
\[\text{restore\_gain}(q) = \mathrm{acc}_q(\text{restored}) - \mathrm{acc}_q(\text{policy}) \in \{-1,0,1\},\]
\[\text{irreversible\_loss}(q)=\max(0,\text{restore\_gain}(q))\cdot\mathbf{1}[G_q \not\subseteq S_P].\]
Here \(\mathrm{acc}_q(c)\in\{0,1\}\) is the judged correctness of \(q\) under condition \(c\in\{\text{restored},\text{policy}\}\). Reader, prompt, decoding (temperature 0), judge, and ranker are held fixed across the policy and restored
conditions, so the comparison varies the availability of \(G_q\) under a common injection procedure;
\(\text{restore\_gain}\) therefore measures the effect of restoring \(G_q\) on correctness, and its positive
part is the accuracy recoverable by read-time restoration. Restoration forces \(G_q\) ahead of ranker
results under a fixed cap, so it can also change which non-gold units appear; A3 bounds the resulting
slack. \(G_q\) comes from the benchmark's evidence labels
over unmodified turn/session units; we do not use a re-represented or derived-summary store because its units would no longer correspond to the labeled evidence.

\textbf{A worked case.} A LongMemEval-S question asks how long the user's daily commute takes. Under an 80k
budget, FIFO has evicted the session that records it, and the reader answers \emph{``I don't know.''}
The restore counterfactual re-injects the evicted session; the reader now answers the correct \emph{``45 minutes
each way''}, so \(\text{restore\_gain}{=}1\) with \(\ge 1\) evicted gold unit --- an irreversible loss whose only fix is
retaining that session. Had the gold session survived in \(S_P\) and the same restoration flipped the answer,
the loss would instead be recoverable; the retriever missed it. The procedure classifies each error in this way,
one question at a time.

\subsection{The recoverability decomposition}\label{the-recoverability-decomposition}

The denominator consists of oracle-answerable policy errors: questions that the reader answers correctly under clean
full-gold injection but incorrectly under the policy. (For evicting policies under forced gold these are
eviction-induced; the no-evict reference and the top-\(k\) regime also admit retrieval-miss/utilization errors,
which the recoverable and residual bins isolate.) Each such error is exactly one of:

\begin{itemize}
\tightlist
\item
  irreversible: restoring \(G_q\) flips wrong\ensuremath{\to}correct and \(\ge 1\) unit of \(G_q\) was evicted (destruction, repaired only by retention);
\item
  recoverable: restoring flips wrong\ensuremath{\to}correct and all of \(G_q\) survived (a retrieval miss, repaired by a better retriever);
\item
  residual: still wrong with \(G_q\) restored (utilization failure, reported separately).
\end{itemize}

We report stacked shares per policy\ensuremath{\times}budget with question-cluster bootstrap CIs and Holm--Bonferroni over the
family. Conditioning on oracle-answerability restricts the decomposition to errors the reader answers correctly
from clean gold, so the bins are not contaminated by intrinsically unanswerable questions.
The residual share separately captures whether the reader could use the evidence.

\subsection{When the bins mean what they are named}\label{when-the-bins-mean-what-they-are-named}

Interpreting the decomposition requires three assumptions. For each, we explain why it is plausible here and how it can fail.

\textbf{A1 (Oracle-answerability).} We score only errors for which the reader answers correctly under clean full-gold injection. With this restriction, a scored error is one the reader could have answered from clean gold, so the bins reflect what the budget did to the evidence rather than the reader's inability to answer. That interpretation can fail only if the clean-gold context is itself unanswerable, which the filter excludes by construction. The same filter also limits what we characterize to the answerable slice of the benchmark, excluding abstention behavior.

\textbf{A2 (Judge validity).} Reader and judge are both GPT-4o-mini. To check grading stability, we calibrated the judge on 40 questions before the main evaluation, obtaining repeat self-consistency of 1.0 against a threshold of \ensuremath{\ge}0.90 (App.~A). This certifies grading stability but not independence from the reader. A correlated reader--judge bias would therefore be invisible to the cross-reader check. As a separate check of grading agreement, an independent GPT-5.5 judge re-grading a stratified sample agrees with GPT-4o-mini at 95.7\% (Cohen's \(\kappa{=}0.90\); App.~C). Both judges are OpenAI models, so a judge from a different provider, or human annotation, would be a stronger check.

\textbf{A3 (Restore tightness).} Full-gold restore is the default; it makes the
irreversible share an upper bound on destruction. Three checks quantify the slack in this upper bound over the 2,276 irreversible cases (App.~C). Restoring only the surviving gold reclassifies 0.3\%. Injecting the complete retained
store (100\% coverage) recovers 2\%, so 98\% of irreversible cases are not latent in retained memory and are consistent with genuine destruction rather than a retrieval miss. Finally, an equal-length placebo (non-gold content at the gold
position) flips 8.3\% against 100\% for the gold restore, so the bin reflects evidence availability, not
placement (8.3\% upper-bounds the combined presentation and unannotated-evidence effect).

\section{Experimental setup}\label{experimental-setup}

The study evaluates four eviction policies at three budgets and under two retrieval regimes on one benchmark,
following a protocol frozen before measurement.

\textbf{Data.} LongMemEval-S \citep{wuLongMemEval2024} contains multi-session histories of approximately 102k tokens,
so all three budgets bind; the dataset version is pinned by the hash reported in App.~A. We use all 470 evidence-labeled questions and exclude abstention
questions, which have no gold location to restore. Evidence session ids contain an \texttt{answer} substring and are never
exposed to the reader or judge.

\textbf{Policies.} We evaluate four eviction policies (FIFO, random, redundancy-aware, and LLM-importance) alongside
a no-evict reference. LLM-importance uses a frozen GPT-4o-mini scorer that assigns each unit a general-importance
score from 1 to 10 without access to the question or gold evidence; it is a simple importance baseline rather
than a state-of-the-art retention system. No policy decision reads gold. We add a
deliberately destructive control (an information-density heuristic that evicts gold at \ensuremath{\approx}3\ensuremath{\times} the baseline
rate); it is the positive control for R3 (\S{}5.3), included to establish that the audit registers destruction when it is present.

\textbf{Budgets.} The three store caps of 8k, 30k, and 80k tokens are all binding for the approximately 102k-token
histories and span severe to mild memory pressure.

\textbf{Retrieval regimes.} Two read-time conditions are evaluated: (a) forced-gold injection, which isolates
destruction by guaranteeing that surviving gold is read; and (b) a frozen top-\(k\) ranker, a realistic condition
in which surviving gold can be missed. The recoverable-share difference, computed as regime (b) minus regime
(a), is itself a result (R2, \S{}5.2).

\textbf{Reader and judge.} Reader and judge are GPT-4o-mini (temperature 0, deterministic); we replicate the
structural results with a stronger GPT-5.4-mini reasoning reader (\S{}5). Scoring is paired per question;
judge variance is on the order of the effect size, so pairing is mandatory. The evaluation code was
construct-validated using a deterministic substitute for the model calls and is released with the per-question
records (link in App.~B). \S3.3 states the validity assumptions for the denominator, judge, and restore.

\textbf{Statistics and protocol.} We run three seeds (deterministic policies use seed 0; App.~A). Under a
pre-specified protocol that was frozen before measurement and included with the artifact, significance is
assessed using a question-cluster bootstrap (10,000 resamples), with Holm--Bonferroni correction within each
hypothesis family. The
pre-specified H1 family is a per-cell
\emph{share \textgreater{} 0} test under forced-gold (regime a). Because forced-gold empties the recoverable bin by
construction, H1 is a construct check that the decomposition is well-defined --- the
substantive R1 magnitude (destruction is the majority, two-bin share \textgreater{} 0.5) is read from the regime-b shares
and their bootstrap CIs (R1). H3 is the regime (b)\ensuremath{-}(a) recoverable delta over the evicting policy\ensuremath{\times}budget
cells. The no-evict cells (zero destruction by construction) are the excluded reference in both families.
For H1/H3, the 12-cell family is the four baseline/control policies (FIFO, random, redundancy-aware, info-density
control) \ensuremath{\times} 3 budgets; the later LLM-importance arm enters Table~\ref{tab:percell} and the H2 matched-accuracy
analysis, not these counts. H2 is the irreversible-rate test over accuracy-matched pairs, amended after the
freeze (R3).

\section{Results}\label{results}

\begin{figure}[t]
\centering
\includegraphics[width=\linewidth]{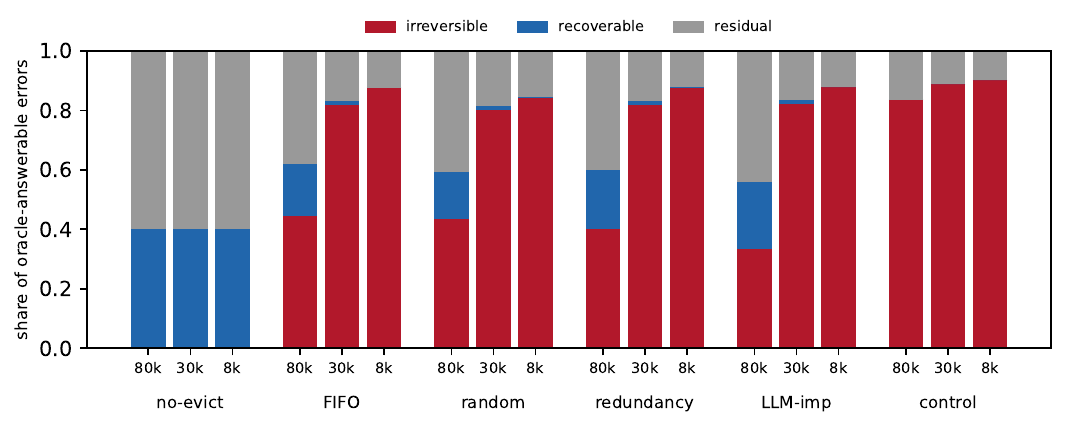}
\caption{The recoverable/irreversible/residual decomposition per policy at three budgets (top-$k$ regime), as
three-bin shares of oracle-answerable policy errors. Destruction dominates on the two-bin denominator, not the
shares plotted here; the no-evict reference has zero irreversible loss by construction. LLM-importance has the lowest irreversible \emph{share}
at 80k, but this is descriptive — at matched accuracy no policy dissociates ($\S$5.3, R3). (The two-bin headline
share is irreversible$/$(irreversible$+$recoverable); per-cell counts in Table~\ref{tab:percell}.)}
\label{fig:decomp}
\end{figure}

\subsection{Destruction is the majority component (R1)}\label{destruction-is-the-majority-component-r1}

Under forced-gold injection, the recoverable bin is empty by construction, so the two-bin share sits at
1.00 across all 12 evicting/control cells. That confirms the
decomposition is well-defined and that forced gold isolates destruction, but it is an algebraic identity
rather than evidence (\S\ref{at-matched-accuracy-no-policy-dissociates-and-the-test-has-power-r3}). The evidence comes from the magnitude under the realistic
top-\(k\) regime. On the two-bin denominator irreversible\(/\)(irreversible+recoverable), the 80k shares are
FIFO 0.71 (CI .61--.81), random 0.73, redundancy-aware 0.67, and control 1.00. Every lower bound exceeds 0.5,
so destruction is the majority component. LLM-importance is the exception: its share of 0.60
(CI .47--.72) straddles 0.5, and no-evict is 0.00 by construction. Table~\ref{tab:percell} reports the per-cell counts; Fig.~\ref{fig:decomp}
plots the smaller three-bin share (\ensuremath{\approx}0.40--0.44 at 80k for FIFO, random, and redundancy-aware).

\begin{table}[t]
\centering\small
\setlength{\tabcolsep}{5pt}
\begin{tabular}{ll rr rrr rr}
\toprule
 & & & & \multicolumn{3}{c}{decomposition} & & \\
\cmidrule(lr){5-7}
policy & budget & $N$ & err & irr & rec & res & two-bin & irr-rate \\
\midrule
no-evict & 80k/30k/8k & 336 & 95 & 0 & 38 & 57 & 0.00 & 0.00 \\
\addlinespace
FIFO & 80k & 336 & 124 & 55 & 22 & 47 & \textbf{0.71} & 0.16 \\
 & 30k & 336 & 243 & 199 & 3 & 41 & 0.99 & 0.59 \\
 & 8k & 336 & 299 & 262 & 0 & 37 & 1.00 & 0.78 \\
\addlinespace
random & 80k & 1008 & 387 & 168 & 61 & 158 & \textbf{0.73} & 0.17 \\
 & 30k & 1008 & 708 & 567 & 11 & 130 & 0.98 & 0.56 \\
 & 8k & 1008 & 894 & 754 & 3 & 137 & 1.00 & 0.75 \\
\addlinespace
redundancy-aware & 80k & 336 & 127 & 51 & 25 & 51 & \textbf{0.67} & 0.15 \\
 & 30k & 336 & 242 & 198 & 3 & 41 & 0.99 & 0.59 \\
 & 8k & 336 & 300 & 263 & 1 & 36 & 1.00 & 0.78 \\
\addlinespace
LLM-importance & 80k & 332 & 102 & 34 & 23 & 45 & \textbf{0.60} & 0.10 \\
 & 30k & 332 & 239 & 196 & 4 & 39 & 0.98 & 0.59 \\
 & 8k & 332 & 302 & 265 & 0 & 37 & 1.00 & 0.80 \\
\addlinespace
control (info-density) & 80k & 336 & 244 & 204 & 0 & 40 & 1.00 & 0.61 \\
 & 30k & 336 & 311 & 276 & 0 & 35 & 1.00 & 0.82 \\
 & 8k & 336 & 327 & 295 & 0 & 32 & 1.00 & 0.88 \\
\bottomrule
\end{tabular}
\caption{\textbf{Eviction loss is destruction-dominated.} At 80k every evicting policy's two-bin share
irr$/$(irr$+$rec) exceeds 0.5 (\textbf{bold}), and it rises to 1.00 as the budget tightens to 8k. Per-cell
counts under the realistic top-$k$ regime (b), primary reader (GPT-4o-mini): the oracle-answerable denominator
$N$, the policy-error count err, the irreversible/recoverable/residual decomposition, the two-bin share, and the
irreversible rate irr$/N$. Regime (a) is the construct reference (rec $\equiv 0$, two-bin $\equiv 1.00$).
\texttt{random} is pooled over 3 seeds ($N{=}1008$); the others are deterministic (seed 0, $N{=}336/332$). The
full grid (both regimes, both readers, bootstrap CIs) and per-question records are in App.~B and the released
artifact.}
\label{tab:percell}
\end{table}

\begin{figure}[t]
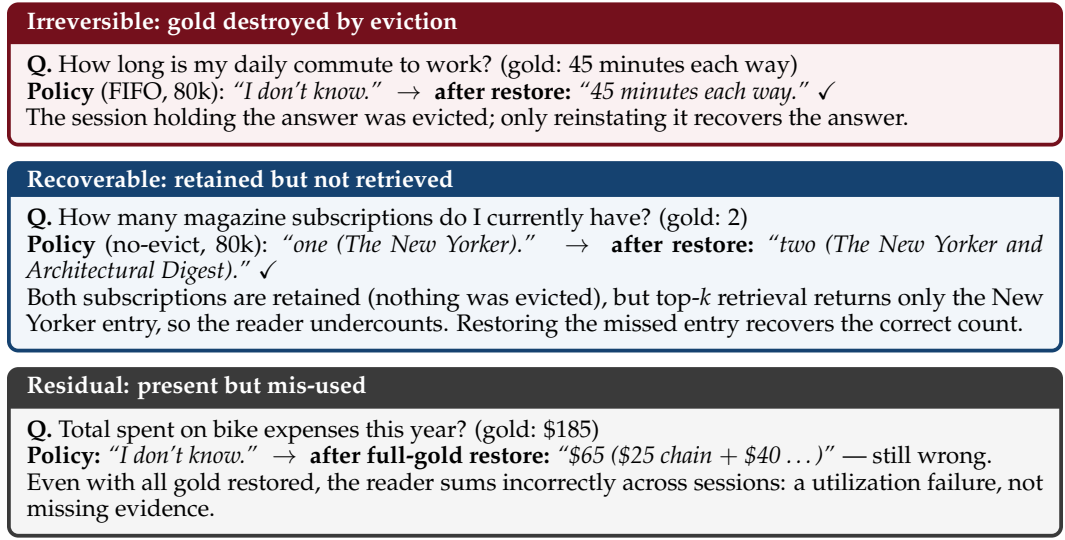

\begin{casebox}{binirr}{Irreversible: gold destroyed by eviction}
\textbf{Q.} How long is my daily commute to work? (gold: 45 minutes each way)\\
\textbf{Policy} (FIFO, 80k): \emph{``I don't know.''} $\;\to\;$ \textbf{after restore:} \emph{``45 minutes each way.''} \checkmark\\
The session holding the answer was evicted; only reinstating it recovers the answer.
\end{casebox}
\begin{casebox}{binrec}{Recoverable: retained but not retrieved}
\textbf{Q.} How many magazine subscriptions do I currently have? (gold: 2)\\
\textbf{Policy} (no-evict, 80k): \emph{``one (The New Yorker).''} $\;\to\;$ \textbf{after restore:} \emph{``two (The New Yorker and Architectural Digest).''} \checkmark\\
Both subscriptions are retained (nothing was evicted), but top-$k$ retrieval returns only the New Yorker entry, so the reader undercounts. Restoring the missed entry recovers the correct count.
\end{casebox}
\begin{casebox}{binres}{Residual: present but mis-used}
\textbf{Q.} Total spent on bike expenses this year? (gold: \$185)\\
\textbf{Policy:} \emph{``I don't know.''} $\;\to\;$ \textbf{after full-gold restore:} \emph{``\$65 (\$25 chain $+$ \$40 \ldots)''} --- still wrong.\\
Even with all gold restored, the reader sums incorrectly across sessions: a utilization failure, not missing evidence.
\end{casebox}
\caption{One real instance of each bin (LongMemEval-S, top-$k$ regime). \textbf{Irreversible}: the gold was
evicted; \textbf{recoverable}: the gold was retained but the retriever missed it; \textbf{residual}: the gold is
present yet the reader still answers wrong. Bin colors match Fig.~\ref{fig:decomp}.}
\label{fig:cases}
\end{figure}

The composition varies with budget and evidence layout. As the budget tightens, the two-bin share climbs
from 0.67--0.73 at 80k for FIFO, random, and redundancy-aware eviction to 1.00 at 8k for all four policies
(Table~\ref{tab:percell}). Evidence layout also matters: single-session
questions are $\sim$95\% irreversible, while multi-session and temporal-reasoning questions have residual shares of 23--34\% and contribute most of the residual errors. Fig.~\ref{fig:cases} gives
one real instance of each bin. With GPT-5.4-mini as the reader, the H1/H3 reject counts and per-policy ordering remain the same,
while the residual mass shrinks (App.~C).

\subsection{The two retrieval regimes are not interchangeable (R2)}\label{the-two-retrieval-regimes-are-not-interchangeable-r2}

Recoverable errors occur under top-\(k\) retrieval but are absent under forced-gold injection by construction. At 80k,
this recoverable-share gap is FIFO +0.29 (CI .19--.39), random +0.27, redundancy-aware +0.33, and no-evict
+1.00, all significant after Holm correction (\(p_\text{adj}{=}.007\)). A small gap for random at 30k is also significant after correction
(\(+0.019\), CI .005--.036, \(p_\text{adj}{=}.042\)). Together, these results give rejections in 4 of 12
evicting cells. The 3 no-evict cells have an empty two-bin denominator under forced gold (irr\({=}\)rec\({=}0\)) and are excluded from that count.

The gap concentrates at 80k because tighter budgets destroy the evidence rather than leave it unretrieved, so fewer recoverable errors remain under top-\(k\) retrieval. The rejection pattern also holds with
GPT-5.4-mini, which reproduces the H3 rejection count of 4-of-12 evicting cells (App.~C). To compare reports at the same budget, researchers must control or at least report the read-time retrieval setting.

\subsection{At matched accuracy, no policy dissociates --- and the test has power (R3)}\label{at-matched-accuracy-no-policy-dissociates-and-the-test-has-power-r3}

R3 is a negative result that requires careful interpretation: the originally specified ``clean null'' was an artifact of evaluating the irreversible share under forced-gold injection. Under forced-gold injection (regime a), the irreversible share is pinned to 1.00 by
construction: all surviving gold is read, so the recoverable bin is empty. That 1.00 contrast therefore cannot
detect a dissociation; it is an algebraic identity, not a test (the protocol was amended after the
freeze to replace this degenerate statistic). H1 and H3 remain as originally specified (confirmatory).
H2 is explicitly flagged as a post-freeze amendment.
Re-cast on the non-degenerate irreversible rate (irr\(/N\), top-\(k\) regime), with pairs counted
as accuracy-matched at \(|\Delta\text{acc}| \le 0.05\) (the pre-specified caliper), no accuracy-matched baseline
pair dissociates at any budget: 0 of 9 (the three baseline pairs \{FIFO/random, FIFO/redundancy-aware,
random/redundancy-aware\} at each of 80k/30k/8k) with \(|\Delta\text{rate}| \le 0.036\) and every CI spanning 0. A
content-aware LLM-importance arm also does not dissociate (0 of 6 accuracy-matched comparisons: versus the three
baselines at 30k and 8k; tests run on the per-pair shared oracle-answerable qids, \(n \approx 329\)). At 80k, LLM-importance has higher accuracy and a lower irreversible rate than the three baselines, but those three comparisons are not accuracy-matched (0.69 vs
\ensuremath{\approx}0.62, beyond the 0.05 caliper) and therefore do not count as H2 dissociations. At 30k and 8k, where accuracy is matched, no irreversible-rate dissociation is detected relative to the baselines.

\textbf{Resolution and positive control.} A failure to reject is informative only if the test could have detected a
real effect. The deliberately destructive control, which evicts gold at \ensuremath{\approx}3\ensuremath{\times} the baseline rate, yields a significant irreversible-rate difference in
all 9 contrasts (\(p<0.001\) under the primary reader; \(p<0.05\) across both readers). These contrasts are not accuracy-matched, so they show the statistic registers a large destructive difference rather than establishing power within the caliper; the matched comparison's sensitivity is the observed paired-bootstrap CI half-width,
1.2--6 percentage points; we therefore report no dissociation \emph{detected at this resolution} rather than statistical
equivalence (an equivalence claim would need a powered TOST, left to a follow-up). The null replicates under
the stronger reader.

\subsection{Residual is reader-utilization failure (R4)}\label{residual-is-reader-utilization-failure-r4}

The residual bin holds errors that survive even the full-gold restore. These cases are oracle-answerable,
and judge strictness accounts for at most 20\%. The remaining explanation is utilization failure rather than
missing evidence: the gold is present, but the reader fails to reason correctly over it, almost always when aggregating information across sessions. In one such case, a \$185 cross-session sum is returned as \emph{``\$65''}.

Under the stronger reader, the residual bin more than halves, from 59\ensuremath{\to}28 for no-evict at 80k with forced-gold injection; the corrected cases are exactly the aggregation errors (App.~C). The residual bin therefore also depends on the reader.

\section{Discussion}\label{discussion}

The measured composition indicates which component needs improvement. Under tight budgets, improving
retention takes priority over improving retrieval (R1): a better retriever offers little benefit until
more evidence is retained. The recoverable component is nonzero but depends on the retrieval regime (R2).
Budget--accuracy frontiers are therefore not directly comparable unless their retrieval regimes are reported.
At matched accuracy, no difference in irreversible rate is detected among the tested policy pairs at a
resolution of 1.2--6 percentage points (R3). The instrument identifies the limiting component but does not select a best policy.

\textbf{Limitations.} The study uses a single benchmark, two readers, and one primary judge; the audit judge shares its provider.
R1/R2 and the R3 null replicate across readers; the residual is reader-dependent.
The audit requires gold labels and an oracle-answerable filter, which limits its use to benchmark analysis.
The eviction arms are policy classes rather than the published systems.
The null is specific to the tested policies and budgets.

\textbf{Ethics.} Here, ``destruction'' refers only to the loss of task evidence through eviction. In deployment,
forgetting may be necessary to meet privacy and data-protection requirements concerning data minimization,
storage limitation, or erasure; in such cases, retaining more information may be undesirable or disallowed.
The instrument quantifies the task cost of a forgetting decision without prescribing whether information
should be retained or deleted; the same decomposition can audit whether a required deletion incurs recoverable
or irreversible task loss.

\section{Conclusion}\label{conclusion}

The restore counterfactual characterizes oracle-answerable policy errors under a memory budget by separating
what eviction destroyed, what retrieval missed, and what the reader failed to use.
These three sources of error require different remedies. For the budgeted-memory literature, this means that budget--accuracy frontiers are not comparable
unless the read-time retrieval regime is reported. Our contribution is an instrument for this analysis; it does not rank policies.

\bibliography{refs}
\bibliographystyle{colm2026_conference}

\appendix
\section{Protocol constants \& reproducibility}\label{protocol-constants-reproducibility}

Reader/judge GPT-4o-mini (temperature 0, deterministic); robustness reader GPT-5.4-mini. Dataset:
LongMemEval-S \texttt{longmemeval\_s\_cleaned.json}, sha256 \texttt{d6f21ea9\ldots{}c3a442} (470 evidence-labeled questions,
$\sim$102k tokens/history, o200k tokenizer); the deprecated original release differs. Store budgets 80k/30k/8k
tokens; read-time top-k = 60; restore inject cap = 2000 tokens (fits every gold set, max \ensuremath{\approx}1,000);
3 seeds (0,1,2; deterministic policies use seed 0). LLM-importance scorer: a frozen GPT-4o-mini rater
(1--10 general importance, never sees the question or gold). Bootstrap: question-cluster, 10,000
resamples, smoothed two-sided p; Holm--Bonferroni within each hypothesis family. Judge calibration (gate run
before the main evaluation, 40 questions): repeat self-consistency 1.0 against a threshold of \ensuremath{\ge}0.90,
oracle-answerable rate 0.95, negative-restore-gain rate 0.00 against a cap of 0.05. Models (OpenAI API): reader/judge
\texttt{gpt-4o-mini}, robustness reader \texttt{gpt-5.4-mini}, importance scorer \texttt{gpt-4o-mini}, and the independent grading-audit
judge \texttt{gpt-5.5} (App.~C); all reproducible through the API. The released artifact pins the model versions
through exact dated snapshot ids and includes all prompts reproduced in App.~D.

\textbf{Retrieval and restore.} The top-\(k\) ranker is a frozen deterministic BM25-lite over the retained store
(lexical overlap; the same ranker for every arm; recency-tiebroken). Regime (b) is pure top-\(k\) (no forced
gold); regime (a) forces surviving gold. A restore adds the missing gold to the store, forces the full gold set
(never truncated), skips duplicate forced ids, then adds ranker results that fit within the remaining inject cap
(2000 tokens); the final injected snippets are ordered chronologically. The question-cluster bootstrap
resamples at the level of the question (all seed-replicates of a sampled question move together), so duplicate
seeds do not inflate the effective sample size. Denominators: the oracle-answerable filter yields \(N{=}336\) for the baseline grid;
the later LLM-importance arm was run as a separate strengthening grid with its own filter (\(N{=}332\)), and every
LLM-vs-baseline paired test runs on the per-pair shared oracle-answerable qids (\(n \approx 329\)). The rate of restores that
turn a correct answer wrong (a negative restore gain) is held below 0.05 by the calibration gate, and such
restores never create a recovery: the irreversible loss is clipped at zero, and only policy-wrong
oracle-answerable errors are binned.

\section{Extended results \& artifact}\label{extended-results-artifact}

Table~\ref{tab:percell} (body) gives the headline per-cell counts under the realistic top-\(k\) regime (b) on the
primary reader. The artifact includes the full grid: both retrieval regimes (a) and (b), both readers
(GPT-4o-mini and GPT-5.4-mini), per-cell bootstrap confidence intervals on every share, and the per-question
records. Regime (a) is the construct reference (recoverable \(\equiv 0\), so the two-bin share \(\equiv 1.00\));
\texttt{random} is pooled over 3 seeds (\(N{=}1008\)), the others deterministic (seed 0, \(N{=}336/332\)).

\textbf{Artifact.} The public release contains the evaluation code (policies, restore, decomposition, and bootstrap),
per-question records for both readers, per-cell tables with bootstrap CIs, the calibration report, and the
figure-generation script: \url{https://github.com/megagonlabs/restore-counterfactual}. The regime-(a) \emph{share \textgreater{} 0}
test is the H1 construct check (recoverable \(\equiv 0\) by construction, so the share \(\equiv 1.00\)); the
no-evict cells (irreversible \(\equiv 0\) by construction) are the excluded zero-destruction reference in the
H1/H3 reject counts (\S{}5).

\section{Reader robustness, residual analysis \& the restore-tightness ablation}\label{reader-robustness-residual-analysis-the-restore-tightness-ablation}

\textbf{Reader robustness (R1/R2).} Replacing the GPT-4o-mini reader with the stronger GPT-5.4-mini reasoning reader
preserves the structural results: identical Holm counts (12 of 12 evicting/control H1 cells reject; 4 of 12
evicting H3 cells reject) and the same per-policy ordering of two-bin shares. The residual bin is the only component that changes.

\textbf{Residual mechanism (R4).} The 1,922 residual records in the primary-reader baseline grid (both retrieval regimes, all budget cells) are
reader-utilization failures: the gold is present,
but the reader fails to reason correctly over it, overwhelmingly in cross-session counting and summation (e.g.~a \$185 total answered as
\emph{``\$65''}). Under a deliberately conservative heuristic, \ensuremath{\le}20\% are possible judge-strictness cases (most are reader
arithmetic errors that merely share entities with the gold); gold-insufficiency is negligible (the cases are
oracle-answerable). The stronger reader resolves these same aggregations (e.g., the \$185
cross-session sum the weaker reader answers \emph{``\$65''} is correct under GPT-5.4-mini), and the bin more than halves
(no-evict@80k: 59\ensuremath{\to}28).

\textbf{Restore-tightness ablations (\S{}3.3, A3).} Three checks bound the irreversible bin's slack, each over the
regime-(b) irreversible cases (seed 0, \(N{=}2{,}276\); 0 API errors each). (i) Surviving gold: re-answering
with only \(G_q \cap S_P\) restored reclassifies 0.3\% (7) as recoverable-in-practice; 87\% had no surviving
gold at all. (ii) Complete retained store: injecting the entire surviving store (all content, not just
annotated gold; mean coverage 1.00, up to the full 80k budget) recovers 2.0\% (45), so 98\% of
irreversible cases remain incorrect even when all retained content is supplied, consistent with genuine
destruction rather than a retrieval-from-retained miss. (iii) Length/position placebo: force-injecting equal-length non-gold content
at the gold position (placebo\(/\)gold token ratio \ensuremath{\le}1.5, near the gold's chronological position) flips 8.3\% (190)
wrong\ensuremath{\to}correct, against 100\% for the gold restore, so the irreversible flips reflect evidence availability, not
placement; the 8.3\% upper-bounds the combined presentation and unannotated-evidence effect (the placebo is
non-annotated, not guaranteed semantically irrelevant). The full-gold restore is therefore a tight upper bound
on destruction.

\textbf{Judge audit (\S{}3.3, A2).} An independent GPT-5.5 judge re-graded a stratified sample of 396 recorded
(question, reference, candidate) items (both the originally incorrect policy answers and the often correct
restored answers) with the frozen judge prompt; 3 GPT-5.5 calls returned errors and were excluded, leaving 393
completed grades spanning both correctness classes (132 correct, 261 incorrect). The GPT-5.5 judge agrees with the
GPT-4o-mini judge at 95.7\% (Cohen's \(\kappa{=}0.90\); on the two-class restored
answers \(\kappa{=}0.83\)), with disagreement dominated by 14 cases where GPT-5.5 is the stricter grader. The
binary grading is thus reliable across models; both judges are OpenAI models, so a judge from a different provider, or human annotation, would be a stronger check.

\section{Prompts}\label{prompts}

Decoding is temperature 0 throughout; \texttt{max\_tokens} = 600 (reader) / 16 (judge) / 4 (importance scorer). Session
ids are never exposed to the reader or judge. The three prompts below are reproduced verbatim
from the released code (\texttt{models.py}); \texttt{\{...\}} marks a field substituted at run time, and \texttt{\{snippets\}} is the
newline-joined rendering of the injected units (or \texttt{(no\ memory\ available)}).

\begin{promptbox}{Reader prompt}
\textbf{System.}\\
You answer a question using ONLY the provided memory snippets from earlier conversations between a user and an assistant. Give the direct answer --- concise but COMPLETE (for a list/order question, include every item in order; for a 'how many days ago' question, compute it from today's date). If the snippets truly do not contain the answer, reply exactly: I don't know.\\[4pt]
\textbf{User.}\\
Today's date is \{question\_date\}.\\
Memory snippets from earlier conversations:\\
\{snippets\}\\[4pt]
Question: \{question\}\\
Answer:
\end{promptbox}

\begin{promptbox}{Judge prompt}
\textbf{System.}\\
Grade whether a candidate answer matches a reference answer for the same question. Reply with exactly one word: CORRECT or INCORRECT. Grade CORRECT if the candidate conveys the same key facts as the reference --- IGNORE articles, capitalization, phrasing, and extra detail. For list/order answers, every reference item must appear (order matters only if the question asks for order). For dates/numbers, the value must match.\\[4pt]
\textbf{User.}\\
Question: \{question\}\\
Reference answer: \{answer\}\\
Candidate answer: \{prediction\}\\[4pt]
Grade (CORRECT or INCORRECT):
\end{promptbox}

\begin{promptbox}{LLM-importance scorer prompt}
\textbf{System.}\\
Rate how generally important this single memory snippet is to remember about the user, on a scale of 1 (trivial small-talk) to 10 (a durable fact, preference, or commitment). Reply with only the integer.\\[4pt]
\textbf{User.}\\
Memory snippet: \{content\}\\
Importance (1-10):
\end{promptbox}

The LLM-importance scorer never sees the question or the gold labels.

\end{document}